\documentclass[
]{ceurart}

\usepackage{listings}

\usepackage{amsmath}
\DeclareMathOperator*{\argmax}{arg\,max}

\begin{document}

\copyrightyear{2025}
\copyrightclause{Copyright for this paper by its authors.
   Use permitted under Creative Commons License Attribution 4.0
   International (CC BY 4.0).}

\conference{TCAI 25: Trustworthy and Collaborative Artificial Intelligence Workshop,
  June 09--11, 2025, Pisa, Italy}
%\conference{}

\title{Task-Agnostic Experts Composition for Continual Learning}

% \tnotetext[1]{You can use this document as the template for preparing your
%   publication. We recommend using the latest version of the ceurart style.}

\author[1]{Luigi Quarantiello}[%
%orcid=0000-0002-0877-7063,
email=luigi.quarantiello@phd.unipi.it
]
\cormark[1]
\address[1]{Computer Science Department, University of Pisa,
  Largo Bruno Pontecorvo 3, 56127, Pisa - Italy}
  
\author[1]{Andrea Cossu}[%
%orcid=0000-0001-7116-9338,
email=andrea.cossu@unipi.it
%url=https://kmitd.github.io/ilaria/,
]

\author[1]{Vincenzo Lomonaco}[%
%orcid=0000-0002-9421-8566,
email=vincenzo.lomonaco@unipi.it
%url=http://conceptbase.sourceforge.net/mjf/,
]

\cortext[1]{Corresponding author.}

\begin{abstract}
Compositionality is one of the fundamental abilities of the human reasoning process, that allows to decompose a complex problem into simpler elements.
Such property is crucial also for neural networks, especially when aiming for a more efficient and sustainable AI framework.
We propose a compositional approach by ensembling \textit{zero-shot} a set of expert models, assessing our methodology using a challenging benchmark, designed to test compositionality capabilities.
We show that our \textit{Expert Composition} method is able to achieve a much higher accuracy than baseline algorithms while requiring less computational resources, hence being more efficient.
\end{abstract}


\maketitle

\section{Introduction}
\begin{comment}
\begin{itemize}
    \item To go towards a more sustainable AI framework, compositionality is one of the main properties.
    \item Compositionality enables knowledge reuse, which reduces the training.
    \item Compositional models are more resilient and robust to distribution shifts, which are common in real world.
    \item In this paper, we explore in particular the composition of knowledge coming from different models
\end{itemize}
\end{comment}
Current AI research has given significant attention on developing large foundation models \cite{foundation_models}. 
Such architectures, typically composed by numerous non-linear layers and billions of parameters, exhibit remarkable results across diverse application domains.
Nonetheless, these achievements come with a substantial computational demand, thereby impacting noticeably on the environment through increased carbon emissions \cite{carbon_emission}.
For this reason, we believe that a change of direction should be promoted, by 
%inverting the trend of the latest years and 
focusing on the development of more efficient and sustainable learning-based solutions.

One possible approach in this sense comes by applying the principle of compositionality.
The standard definition of compositionality \cite{compositionality_definition} reads as follows:
\begin{displayquote}
\emph{The meaning of a compound expression is a function of the meanings of its parts and of the way they are syntactically combined.}
\end{displayquote}
In other words,
%the compositionality property
it refers to the ability of recognizing the whole as the sum of its parts.
%This concept has been studied for more than 30 years \cite{modularity_biologically, modularity_nn}. In these preliminary studies, the authors took inspiration from biological properties of the human brain, which appears to function in an highly compositional way.
This concept has been studied in the context of AI for more than 30 years \cite{modularity_biologically, modularity_nn}, starting by taking inspiration from biological properties of the human brain, which appears to function in a highly compositional way.
In computer vision applications, in particular, it can be seen as the capacity of classifying an image by detecting the objects present in the scene.

One of the main advantages that comes from applying the compositionality property is the one of knowledge reuse.
In fact, one could have a set of pretrained models, each able to recognize particular patterns or objects, and reuse them for different applications, selecting only the needed networks and composing their answers.
Such composition could be \textit{zero-shot}, \textit{i.e.} directly applying the models without further training, or \textit{few-shot}, allowing for a short training procedure with only a small set of samples.

%As it is clear,
Following this approach leads to a much more sustainable AI framework, since it allows to effectively exploit the knowledge acquired by the models, reusing them multiple times, while reducing the need for long and expensive training processes.
Compositional models, given that they focus on small sub-components, are also \textit{by design} invariant to larger scale changes, such as variations in the background of an image or in the relations among objects.
%Such models are also more versatile, being able to fit to different applications and data distributions by simply adjusting the models composition.
%Therefore, these architecture are much more robust to shifts in the data than ``\textit{monolithic}'' ones, which are very common in real-world applications.
Additionally, by simply adjusting the models composition, such architectures are able to fit seamlessly to different applications and data distributions; these makes them much more versatile and robust than their ``\textit{monolithic}'' counterparts.

In this work, we mainly explore the composition of knowledge coming from different expert models.
We will show that, by enforcing compositionality, we are able to largely surpass the performance of known approaches, using a compositional benchmark to asses our method.

\section{Experts Composition}
\begin{comment}
\begin{itemize}
    \item Our objective is to reuse and compose the knowledge of a set of pretrained experts.
    \item We employ the CGQA benchmark from \cite{cfst}; description of the CR\_GQA and CGQA datasets.
\end{itemize}
\end{comment}

To show the benefits of enforcing compositionality to neural models, our proposed methodology consists mainly in two steps: first, the training of expert models, and then their application in a compositional scenario.
Ideally, especially considering that the number of pretrained models available online is always increasing, it could be possible to simply download the needed networks, thus avoiding the training process at all.
Unfortunately, in the case of our work, we could not find adequate \textit{ready-to-use} models. 
Hence, we opted to train our own networks, nonetheless keeping them as small as possible to maintain an efficient approach.
%In fact, we selected the ResNet-18 \cite{resnet} architecture, which is the smallest version of a well-known, state-of-the-art Computer Vision model.

\begin{figure}
    \centering
    \includegraphics[width=0.9\linewidth]{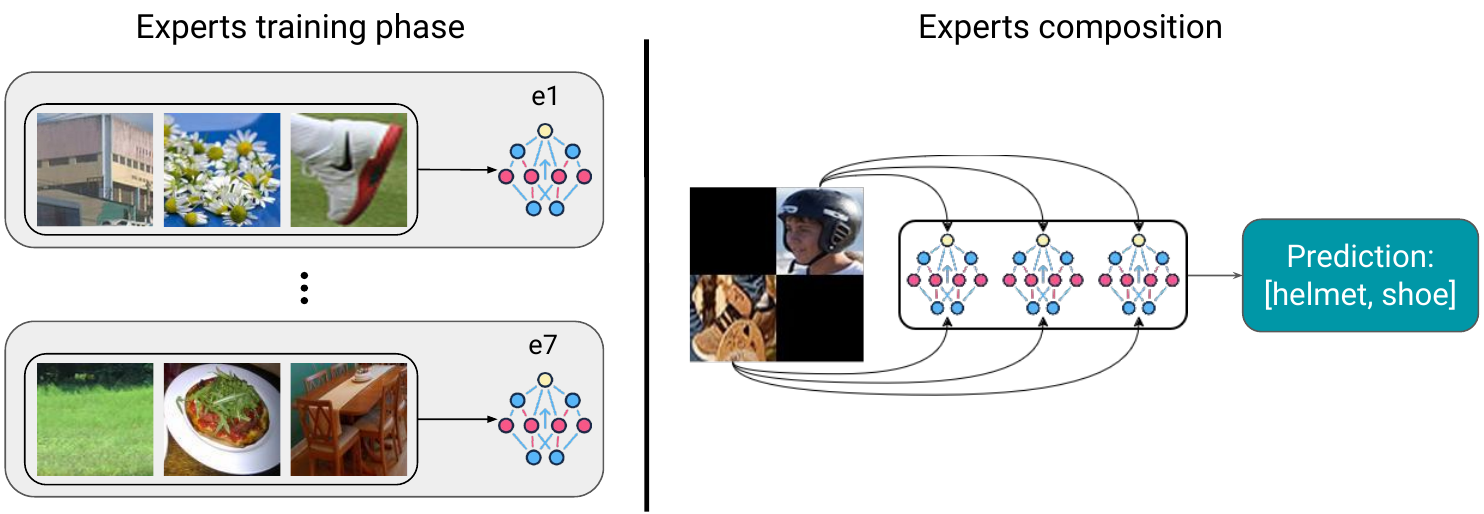}
    %\caption{On the left, the training of the expert models, each on a different subset of cropped images from GQA. On the right, the experts composition is tested on CGQA.}
    \caption{Left: training of the expert models, each on a different subset of cropped images from GQA. Right: the experts composition is tested on CGQA.}
    \label{fig:experts_composition}
\end{figure}

To train the experts, we employed cropped images from the GQA \cite{GQA} dataset.
%GQA is a real-world, large-scale visual question-answering dataset, that offers also a rich set of information about the objects in the images.
We selected 21 objects
%from the ones present
from the dataset, we used the information about their bounding boxes to crop the original images and we resized them to a resolution of 98x98 pixels (Figure \ref{fig:experts_composition}, left).
%After this preprocessing phase, the resulting dataset is divided in 200000 training samples, 10000 validation samples and 20000 test images.
%An example of the training images is provided in Figure \ref{fig:cr_gqa}.
\begin{comment}
\begin{figure}
\centering
\begin{minipage}{.5\textwidth}
  \centering
  \includegraphics[width=0.4\linewidth]{img/105804.jpg}
  \includegraphics[width=0.4\linewidth]{img/104657.eps}
  \centering
  \captionof{figure}{Cropped images from GQA.}
  \label{fig:cr_gqa}
\end{minipage}%
\begin{minipage}{.5\textwidth}
  \centering
  \includegraphics[width=0.4\linewidth]{img/10000.eps}
  \includegraphics[width=0.4\linewidth]{img/100061.eps}
  \captionof{figure}{Images from CGQA.}
  \label{fig:cgqa}
\end{minipage}
\end{figure}
\end{comment}
We assigned 3 unique classes to each model, thus resulting in 7 experts.
Each model is trained to specialize on its classes, while a generic ``\textit{other}'' class label is used as a container of the remainder of classes.
%To allow models to learn a certain discriminate power, we employed a ``1 vs All'' classification paradigm: each model was trained on its corresponding classes, plus an ``\textit{other}'' class, which was assigned to samples with other labels.
%In this way, an expert could also detect when an image is out of its training data distribution.
In this way, an expert can both provide an answer when prompted with samples from its classes, but also detect when an image is out of its scope.

These trained models were then tested on the highly compositional CGQA benchmark \cite{cgqa}, that contains samples made by a 2x2 grid, in which 2 cropped object images taken from GQA are inserted (Figure \ref{fig:experts_composition}, right).
%As reported in Figure \ref{fig:cgqa}, this dataset contains samples made by a 2x2 grid, in which 2 cropped object images taken from GQA are inserted.
%The data is then divided into several experiences, each with a subset of the classes, to allow for a continual training procedure.
%This benchmark is used by the authors to assess a set of four baseline algorithms.
CGQA was originally designed for Continual Learning \cite{continual_learning}, a Machine Learning paradigm in which models are trained on streams of different tasks in dynamic environments; nonetheless, the benchmark can be easily adapted for offline testing.
In the paper, two main scenarios are presented, the task-incremental and class-incremental learning cases.
We will compare our method against the second one, which is the most challenging setting, since no task identifier is provided and a single-head classifier must be employed.
%We will focus in particular on the class-incremental learning case, the setting in which no task identifier is provided, therefore a single-head classifier is employed.

%To test the compositionality capabilities of our approach, we divide each query image into its quadrants and we provide them to the set of experts.
%We then retrieve the prediction with the highest confidence for each quadrant, and we create the compositional label accordingly.

Our approach can be formalized as follows: for each query image, we extract its composing quadrants; then, the prediction for each quadrant is computed as:
$$
\forall \; \text{expert}, \quad p = \argmax \;\text{expert}(q)\\
$$
where $q$ is a quadrant from an image.
Lastly, the entire prediction for a compositional image is given by the concatenation of the predictions on its quadrants:
$$p_\text{comp} = p_1 \cup p_2$$
In this fashion, no additional training procedure is needed to recognize the objects composition, since the experts are employed \textit{zero-shot}.
This results in a substantial saving in training effort with respect to related approaches, which instead need to be fine-tuned for each experience, incurring also in loss of accuracy due to the catastrophic forgetting phenomenon.

%\textcolor{red}{
With the same models, we also explored the use of a finetuned approach to enforce compositionality.
Specifically, we used the few-shot \textit{sys} stream provided by CGQA, defining 300 different tasks, called experiences.
For each experience, we select the top-performing experts, we freeze them and we train a single-layer linear classifier on top of their concatenated features.
To determine which experts to use, we perform a forward pass over all models using the training data, collecting the sum of the top logits per model.
This value represents
%the excitement level of a given expert, \textit{i.e.}
how much current data are \textit{in-distribution} with the training data of a given model.

\section{Experimental Results}
\begin{comment}
\begin{itemize}
    \item Experimental setup: model used, important hyperparameters.
    \item Show the results --- our model is better than any other baseline, even when using pretrained backbones.
    \item Stress the efficiency --- our approach does not need any further training, which results in less training time and resource consumption, and it is able to achieve better results.
\end{itemize}
\end{comment}

%\vspace{-1mm}
\subsection{Training of the Experts}
%\vspace{-1mm}
%In this section, we will show the experimental results we obtained with our compositional approach.
Firstly, we trained a set of neural networks to define the expert models.
Our main objective was to design an efficient and sustainable method; therefore, we selected the ResNet-18 architecture \cite{resnet}, the smallest version of the well-known, \textit{state-of-the-art} computer vision model.
As discussed in the previous section, we trained our models on images of objects from the GQA dataset.\\
The resulting dataset contains 21 different classes; on average, each class has about 11.000 samples.
It was then divided so that each expert is trained to specialize on 3 classes, thus resulting in a set of 7 models.
%For the training of the expert models, we followed the dataset splitting suggested by the authors of GQA.
Following the dataset splitting
%suggested by the authors
of GQA, on average,
%Hence, on average,
%following the \textit{1 vs all} paradigm,
each ResNet was trained on about 38.000 images, with about 2000 images as validation set, and 3800 test images.

\begin{figure}
\centering
\begin{minipage}{.4\textwidth}
  \vspace{-7mm}
  \centering
  \includegraphics[width=0.88\linewidth]{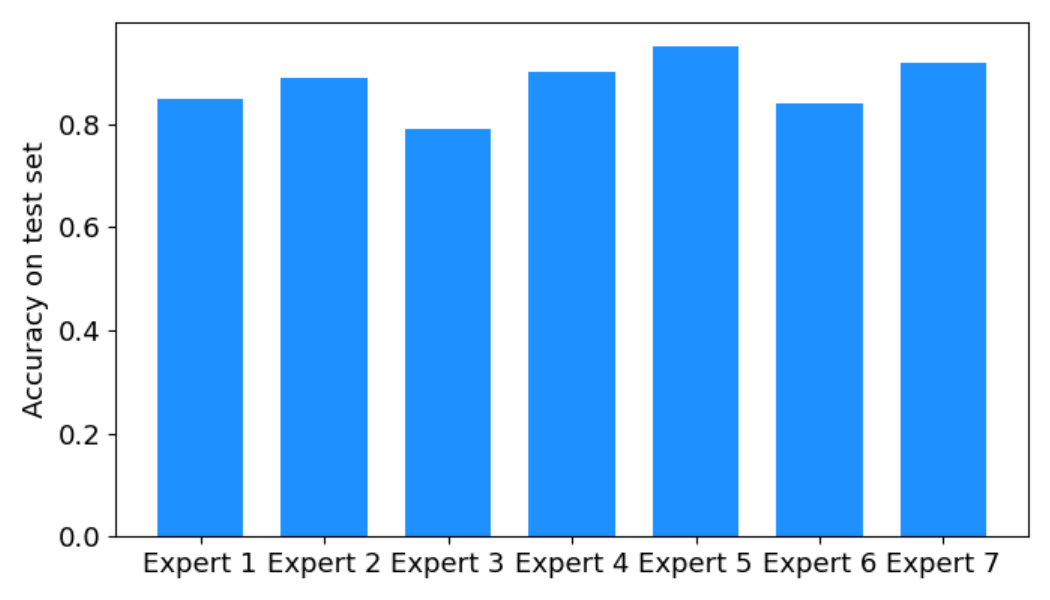}
  \centering
  \captionof{figure}{The accuracy of the\\expert models.}
  \label{fig:experts_training_results}
\end{minipage}%
\begin{minipage}{.6\textwidth}
  \centering
  \includegraphics[width=0.86\linewidth]{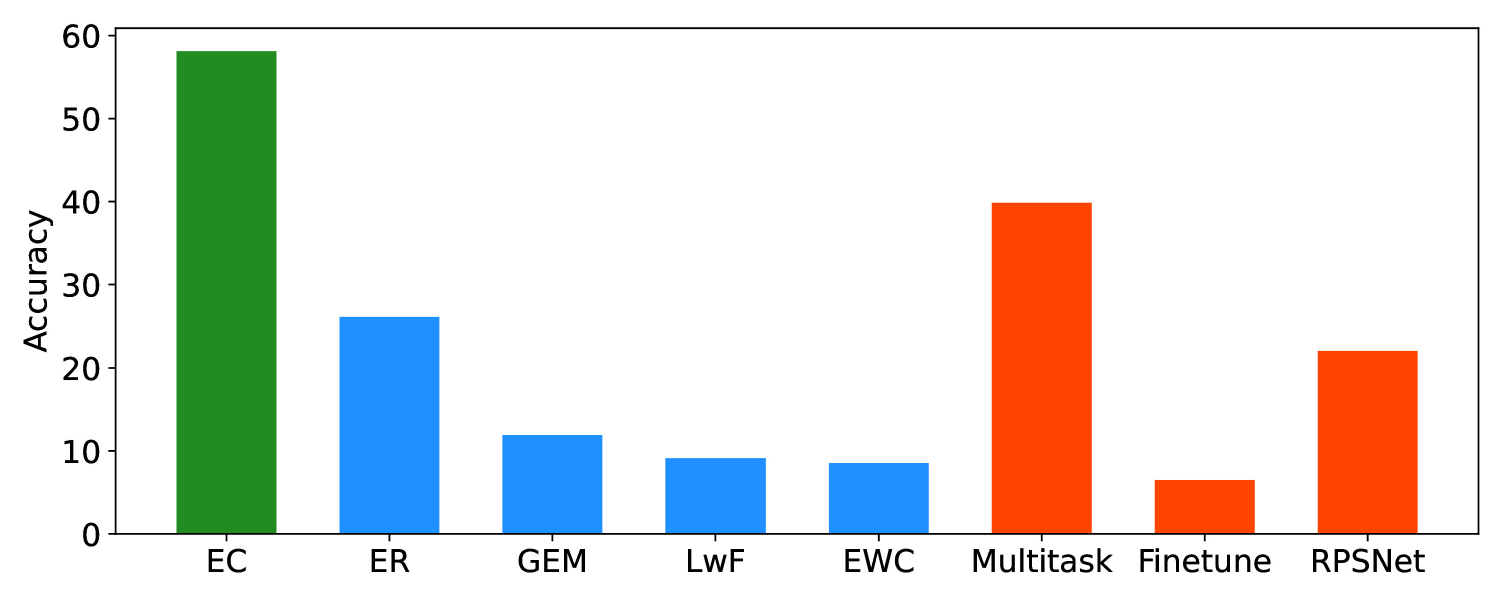}
  \captionof{figure}{Accuracy of our method Experts Composition (EC)\\\textit{wrt} baselines. In blue, CL methods using a pretrained backbone.\\In red, results taken from \cite{cgqa}.}
  \label{fig:comp_results}
\end{minipage}
\end{figure}

In Figure \ref{fig:experts_training_results}, the accuracy of these models on their respective test set is plotted.
The employed architectures have 18 layers, with 3x3 convolutional kernels; we used Adam as optimizer, with a learning rate of 1e-3.
%As it is clear from the plot,
The resulting models have a similar behavior, with an average accuracy of 88\%.
%On the other hand, for what concerns the computational demand,
Conversely, in terms of computational demand,
%we performed the training procedures using
we used a NVIDIA Tesla T4 GPU, taking about 13 minutes for each training loop.
Therefore, we were able to achieve optimal performance over a large set of images, while using few resources and in a short time.
%Such efficiency will be further increased during the testing on CGQA, by enforcing compositionality and knowledge reuse, as discussed in the next section.

%\vspace{-1mm}
\subsection{Experts Composition}
%\vspace{-1mm}
Successively, we tested the composition of our pretrained experts on the CGQA benchmark.
%a benchmark specifically designed to test compositionality capabilities of neural models.
%CGQA is a benchmark specifically designed to test compositionality capabilities of neural models; it was originally designed to train Continual Learning models, but it can be easily adapted for an offline testing.
In its original paper, the authors used the dataset with Continual Learning techniques, namely ER \cite{er}, EWC \cite{ewc}, LwF \cite{lwf}, GEM \cite{gem} and RPSnet \cite{rpsnet}, as well as the \textit{Multitask} and \textit{Finetune} baselines.\\
For our experiments, we used the data from the \textit{con} stream, which contains combinations of objects from the 21 different classes.
The dataset contains 100 different combinations, \textit{i.e.} labels, with 100 test instances for each combination.

In Figure \ref{fig:comp_results}, we report the results we obtained with our EC method, together with the ones from the baseline approaches.
%In particular, we present the accuracy of the \textit{zero-shot} composition, retrieving the label for each quadrant of the images from the set of experts.
%For the baseline algorithms, to compare the approaches more fairly, we reproduced the experimental setting from the original CGQA paper, but starting from a pretrained model.
%Instead of using randomly initialized networks, as it is done in the original work, we pretrained a ResNet on the entire dataset extracted from GQA, designing an expert across all 21 object classes.
For some of the baselines, to compare the approaches more fairly, instead of using randomly initialized networks, we pretrained a ResNet on the entire dataset extracted from GQA, designing a single expert across all 21 object classes.
For the other approaches, we took the accuracy results directly from the original paper.
%In this way, both our method and the other algorithms employ networks pretrained on the same images.\\
%Moreover, instead of starting from randomly initialized networks, we pretrained a ResNet model on the entire dataset extracted from GQA, designing an expert across all 21 object classes.
\\As it is clear from the plot, our approach abundantly surpass all the other algorithms, obtaining an accuracy over the test set of 58\%.
A second important observation is that, being an un-trained composition, the performance of our approach does not change over time, and it is already optimal from the first experience.
Conversely, the other methods need to be trained across the experiences, being able to reach their top accuracy only at the end.
For these experiments, we used the Avalanche library \cite{avalanche}.\\
The best behavior is given by the ER algorithm, which achieves 26\% of accuracy; in addition, on average, each of the four baselines we reproduced took 3 hours and 45 minutes to complete the training, while our experts do not require any additional fine-tuning phase.
%Furthermore, our experts do not require any additional training phase, while the other methods need to be finetuned when new objects compositions arrive.
In other words, our approach turns out to be more efficient than the other ones from a computational viewpoint, but also more robust to shifts in the data distribution, which are common especially in real-world scenarios.

%\textcolor{red}{
We also tried to compose the expert models using a custom classification head, trained using few samples for each class.
We set the number of experiences to 300, as in the CGQA paper, and took the average accuracy over the test set.
We experimented the composition of 3, 5 and 7 experts, reporting the obtained results in Table \ref{tab:finetuned_comp_results}.
%}
%\textcolor{red}{
%The results of this approach are reported in Table \ref{tab:finetuned_comp_results}.
Such method has much worse performance, especially when employing few experts.
This is probably due to the fact that the training samples are too few to train an effective classifier.
We plan to investigate deeper the problem, since we believe
%that zero-shot composition could not be always applicable, and
that,
%there are settings in which
in some setting, zero-shot composition is not adequate and a certain degree of fine-tuning is needed to adapt to the problem.
%}

\begin{table}[h!]
  \centering
  \scalebox{1}{
  \begin{tabular}{ccc}
    \toprule
    Number of Experts & Avg. Test Set Accuracy (\%)\\
    \midrule
    3& 18.82 $\pm$ 0.98\\
    5& 35.77 $\pm$ 1.68\\
    7& 42.82 $\pm$ 1.45\\
    \bottomrule
  \end{tabular}}
  \caption{Results from the fine-tuned expert composition.}\label{tab:finetuned_comp_results}
\end{table}

\section{Conclusion}
In this work, we presented a study about enforcing compositionality in neural models.
Such property plays a crucial role in the context of a sustainable AI framework, since it allows to reuse pretrained models for different applications, saving the computational resources needed for additional training procedure.
We trained different expert models on a set of objects taken from the GQA dataset, and then tested their composition on CGQA, an highly compositional benchmark.
We assessed our methodology against several baseline approaches, obtaining better results both in terms of performance and efficiency.
Specifically, we managed to nearly double the accuracy achieved by other methods, while consuming much less computational resources and training time.

For this work, we employed a challenging benchmark for compositional approaches, but it is nonetheless a synthetic and artificial dataset.
As future work, we plan to expand our method to more realistic and complex settings.\\
Moreover, we consider this work as an initial step, and we believe that additional endeavours can be spent to deeper explore such approaches, in different directions.
As an example, it could be interesting to study how to obtain expert systems from pretrained architectures, to further enhance the efficiency of this kind of solutions.
Lastly, we want to experiment more on fine-tuned compositions, to adapt the method for settings in which zero-shot composition is not enough.

\bibliography{ref}

\begin{thebibliography}{15}
\expandafter\ifx\csname natexlab\endcsname\relax\def\natexlab#1{#1}\fi
\providecommand{\url}[1]{\texttt{#1}}
\providecommand{\href}[2]{#2}
\providecommand{\path}[1]{#1}
\providecommand{\DOIprefix}{doi:}
\providecommand{\ArXivprefix}{arXiv:}
\providecommand{\URLprefix}{URL: }
\providecommand{\Pubmedprefix}{pmid:}
\providecommand{\doi}[1]{\href{http://dx.doi.org/#1}{\path{#1}}}
\providecommand{\Pubmed}[1]{\href{pmid:#1}{\path{#1}}}
\providecommand{\bibinfo}[2]{#2}
\ifx\xfnm\relax \def\xfnm[#1]{\unskip,\space#1}\fi
%Type = Article
\bibitem[{Bommasani et~al.(2021)Bommasani, Hudson, Adeli, Altman, Arora, von Arx, Bernstein, Bohg, Bosselut, Brunskill et~al.}]{foundation_models}
\bibinfo{author}{R.~Bommasani}, \bibinfo{author}{D.~Hudson}, \bibinfo{author}{E.~Adeli}, \bibinfo{author}{R.~Altman}, \bibinfo{author}{S.~Arora}, \bibinfo{author}{S.~von Arx}, \bibinfo{author}{M.~Bernstein}, \bibinfo{author}{J.~Bohg}, \bibinfo{author}{A.~Bosselut}, \bibinfo{author}{E.~Brunskill}, et~al.,
\newblock \bibinfo{title}{On the opportunities and risks of foundation models},
\newblock \bibinfo{journal}{arXiv preprint arXiv:2108.07258}  (\bibinfo{year}{2021}).
%Type = Article
\bibitem[{Patterson et~al.(2021)Patterson, Gonzalez, Le, Liang, Munguia, Rothchild, So, Texier, and Dean}]{carbon_emission}
\bibinfo{author}{D.~Patterson}, \bibinfo{author}{J.~Gonzalez}, \bibinfo{author}{Q.~Le}, \bibinfo{author}{C.~Liang}, \bibinfo{author}{L.~Munguia}, \bibinfo{author}{D.~Rothchild}, \bibinfo{author}{D.~So}, \bibinfo{author}{M.~Texier}, \bibinfo{author}{J.~Dean},
\newblock \bibinfo{title}{Carbon emissions and large neural network training},
\newblock \bibinfo{journal}{arXiv preprint arXiv:2104.10350}  (\bibinfo{year}{2021}).
%Type = Book
\bibitem[{Werning et~al.(2012)Werning, Hinzen, and Machery}]{compositionality_definition}
\bibinfo{author}{M.~Werning}, \bibinfo{author}{W.~Hinzen}, \bibinfo{author}{E.~Machery}, \bibinfo{title}{The Oxford handbook of compositionality}, \bibinfo{publisher}{OUP Oxford}, \bibinfo{year}{2012}.
%Type = Article
\bibitem[{Braham and Hamblen(1990)}]{modularity_biologically}
\bibinfo{author}{R.~Braham}, \bibinfo{author}{J.~Hamblen},
\newblock \bibinfo{title}{The design of a neural network with a biologically motivated architecture},
\newblock \bibinfo{journal}{IEEE transactions on neural networks} \bibinfo{volume}{1} (\bibinfo{year}{1990}) \bibinfo{pages}{251--262}.
%Type = Article
\bibitem[{Hussain(1995)}]{modularity_nn}
\bibinfo{author}{T.~Hussain},
\newblock \bibinfo{title}{Modularity within neural networks},
\newblock \bibinfo{journal}{Queens University Kingston, Ontario, Canada}  (\bibinfo{year}{1995}).
%Type = Inproceedings
\bibitem[{Hudson and Manning(2019)}]{GQA}
\bibinfo{author}{D.~Hudson}, \bibinfo{author}{C.~Manning},
\newblock \bibinfo{title}{Gqa: A new dataset for real-world visual reasoning and compositional question answering},
\newblock in: \bibinfo{booktitle}{Proceedings of the IEEE/CVF conference on computer vision and pattern recognition}, \bibinfo{year}{2019}, pp. \bibinfo{pages}{6700--6709}.
%Type = Article
\bibitem[{Liao et~al.(2024)Liao, Wei, Jiang, Zhang, and Ishibuchi}]{cgqa}
\bibinfo{author}{W.~Liao}, \bibinfo{author}{Y.~Wei}, \bibinfo{author}{M.~Jiang}, \bibinfo{author}{Q.~Zhang}, \bibinfo{author}{H.~Ishibuchi},
\newblock \bibinfo{title}{Does continual learning meet compositionality? new benchmarks and an evaluation framework},
\newblock \bibinfo{journal}{Advances in Neural Information Processing Systems} \bibinfo{volume}{36} (\bibinfo{year}{2024}).
%Type = Article
\bibitem[{Parisi et~al.(2019)Parisi, Kemker, Part, Kanan, and Wermter}]{continual_learning}
\bibinfo{author}{G.~Parisi}, \bibinfo{author}{R.~Kemker}, \bibinfo{author}{J.~Part}, \bibinfo{author}{C.~Kanan}, \bibinfo{author}{S.~Wermter},
\newblock \bibinfo{title}{Continual lifelong learning with neural networks: A review},
\newblock \bibinfo{journal}{Neural networks} \bibinfo{volume}{113} (\bibinfo{year}{2019}) \bibinfo{pages}{54--71}.
%Type = Inproceedings
\bibitem[{He et~al.(2016)He, Zhang, Ren, and Sun}]{resnet}
\bibinfo{author}{K.~He}, \bibinfo{author}{X.~Zhang}, \bibinfo{author}{S.~Ren}, \bibinfo{author}{J.~Sun},
\newblock \bibinfo{title}{Deep residual learning for image recognition},
\newblock in: \bibinfo{booktitle}{Proceedings of the IEEE conference on computer vision and pattern recognition}, \bibinfo{year}{2016}, pp. \bibinfo{pages}{770--778}.
%Type = Article
\bibitem[{Chaudhry et~al.(2019)Chaudhry, Rohrbach, Elhoseiny, Ajanthan, Dokania, Torr, and Ranzato}]{er}
\bibinfo{author}{A.~Chaudhry}, \bibinfo{author}{M.~Rohrbach}, \bibinfo{author}{M.~Elhoseiny}, \bibinfo{author}{T.~Ajanthan}, \bibinfo{author}{P.~Dokania}, \bibinfo{author}{P.~Torr}, \bibinfo{author}{M.~Ranzato},
\newblock \bibinfo{title}{On tiny episodic memories in continual learning},
\newblock \bibinfo{journal}{arXiv preprint arXiv:1902.10486}  (\bibinfo{year}{2019}).
%Type = Article
\bibitem[{Kirkpatrick et~al.(2017)Kirkpatrick, Pascanu, Rabinowitz, Veness, Desjardins, Rusu, Milan, Quan, Ramalho, Grabska-Barwinska et~al.}]{ewc}
\bibinfo{author}{J.~Kirkpatrick}, \bibinfo{author}{R.~Pascanu}, \bibinfo{author}{N.~Rabinowitz}, \bibinfo{author}{J.~Veness}, \bibinfo{author}{G.~Desjardins}, \bibinfo{author}{A.~Rusu}, \bibinfo{author}{K.~Milan}, \bibinfo{author}{J.~Quan}, \bibinfo{author}{T.~Ramalho}, \bibinfo{author}{A.~Grabska-Barwinska}, et~al.,
\newblock \bibinfo{title}{Overcoming catastrophic forgetting in neural networks},
\newblock \bibinfo{journal}{Proceedings of the national academy of sciences} \bibinfo{volume}{114} (\bibinfo{year}{2017}) \bibinfo{pages}{3521--3526}.
%Type = Article
\bibitem[{Li and Hoiem(2017)}]{lwf}
\bibinfo{author}{Z.~Li}, \bibinfo{author}{D.~Hoiem},
\newblock \bibinfo{title}{Learning without forgetting},
\newblock \bibinfo{journal}{IEEE transactions on pattern analysis and machine intelligence} \bibinfo{volume}{40} (\bibinfo{year}{2017}) \bibinfo{pages}{2935--2947}.
%Type = Article
\bibitem[{Lopez-Paz and Ranzato(2017)}]{gem}
\bibinfo{author}{D.~Lopez-Paz}, \bibinfo{author}{M.~Ranzato},
\newblock \bibinfo{title}{Gradient episodic memory for continual learning},
\newblock \bibinfo{journal}{Advances in neural information processing systems} \bibinfo{volume}{30} (\bibinfo{year}{2017}).
%Type = Article
\bibitem[{Rajasegaran et~al.(2019)Rajasegaran, Hayat, Khan, Khan, and Shao}]{rpsnet}
\bibinfo{author}{J.~Rajasegaran}, \bibinfo{author}{M.~Hayat}, \bibinfo{author}{S.~Khan}, \bibinfo{author}{F.~Khan}, \bibinfo{author}{L.~Shao},
\newblock \bibinfo{title}{Random path selection for continual learning},
\newblock \bibinfo{journal}{Advances in neural information processing systems} \bibinfo{volume}{32} (\bibinfo{year}{2019}).
%Type = Inproceedings
\bibitem[{Lomonaco et~al.(2021)Lomonaco, Pellegrini, Cossu, Carta, Graffieti, Hayes, De~Lange, Masana, Pomponi, Van~de Ven et~al.}]{avalanche}
\bibinfo{author}{V.~Lomonaco}, \bibinfo{author}{L.~Pellegrini}, \bibinfo{author}{A.~Cossu}, \bibinfo{author}{A.~Carta}, \bibinfo{author}{G.~Graffieti}, \bibinfo{author}{T.~Hayes}, \bibinfo{author}{M.~De~Lange}, \bibinfo{author}{M.~Masana}, \bibinfo{author}{J.~Pomponi}, \bibinfo{author}{G.~Van~de Ven}, et~al.,
\newblock \bibinfo{title}{Avalanche: an end-to-end library for continual learning},
\newblock in: \bibinfo{booktitle}{Proceedings of the IEEE/CVF Conference on Computer Vision and Pattern Recognition}, \bibinfo{year}{2021}, pp. \bibinfo{pages}{3600--3610}.

\end{thebibliography}

\end{document}